\documentclass[10pt,twocolumn,letterpaper]{article}

\usepackage{wacv}
\usepackage{times}
\usepackage{epsfig}
\usepackage{graphicx}
\usepackage{amsmath}
\usepackage{amssymb}

\usepackage[skip=0pt]{caption}
\usepackage{booktabs}
\usepackage{multirow}
\usepackage{ctable}

\usepackage{subcaption}

\usepackage[pagebackref=true,breaklinks=true,letterpaper=true,colorlinks,bookmarks=false]{hyperref}

\wacvfinalcopy 

\def\wacvPaperID{697} 

\ifwacvfinal\fi
\begin{document}
 \thispagestyle{empty}
\title{High Accuracy Face Geometry Capture using a Smartphone Video}

\author{Shubham Agrawal\\
Carnegie Mellon University\\
{\tt\small sagrawa1@alumni.cmu.edu}
\and
Anuj Pahuja\\
Carnegie Mellon University\\
{\tt\small apahuja@andrew.cmu.edu}
\and
Simon Lucey\\
Carnegie Mellon University\\
{\tt\small slucey@andrew.cmu.edu}
}


\twocolumn[{%
\renewcommand\twocolumn[1][]{#1}%
\vspace{-1em}
\maketitle
\vspace{-1em}
\begin{center}
   \centering \includegraphics[width=\textwidth]{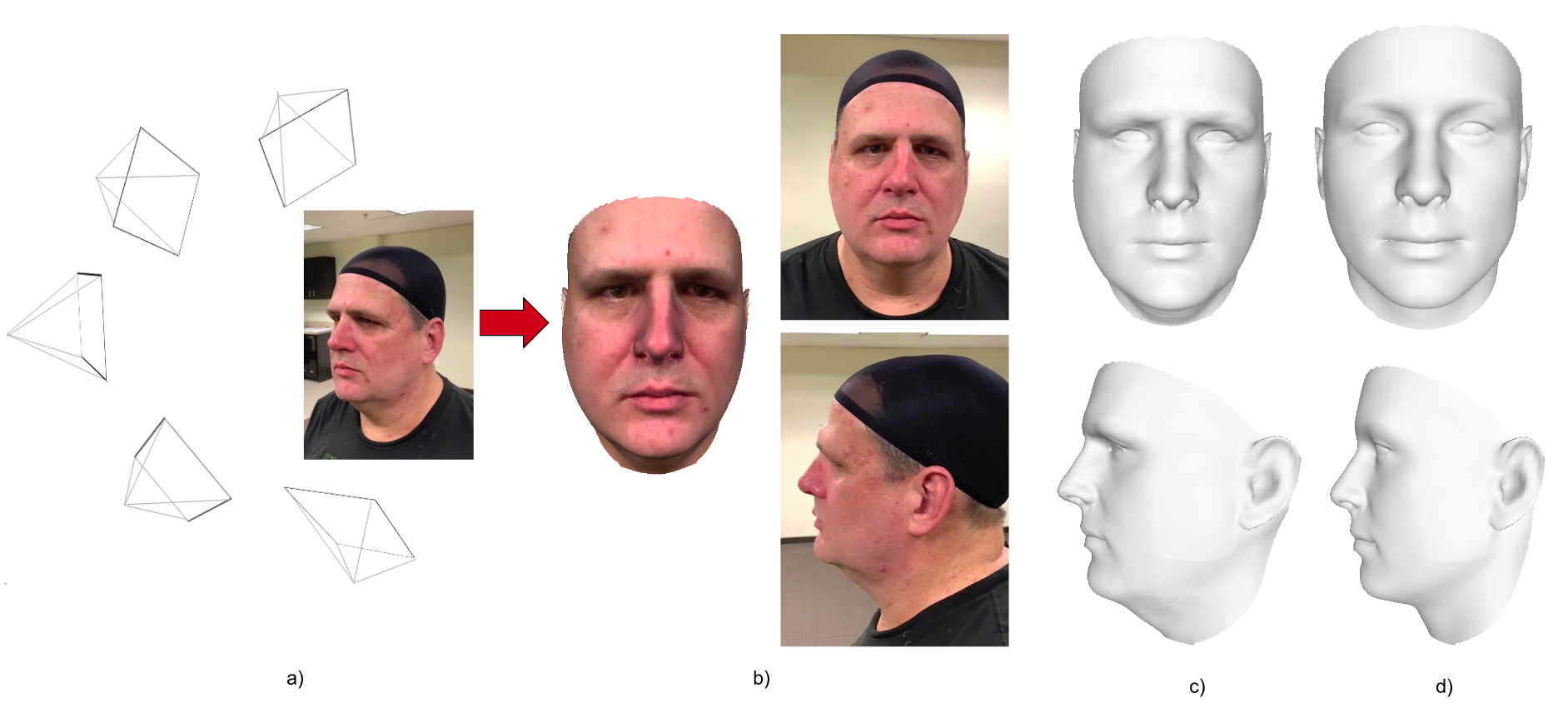} \captionof{figure}{a) Our method uses a single video clip of a subject taken in an unconstrained environment to reconstruct a highly accurate mesh of the face geometry. b) Our textured mesh side-by-side with input images c) Our method (left) compared to d) the state-of-the-art method of \cite{hernandez2017accurate} (right), front and profile view.}
   \label{cover}
\end{center}%
}]

\begin{abstract}
 \vspace{-10pt}
    What's the most accurate 3D model of your face you can obtain while sitting at your desk? We attempt to answer this question in our work. High fidelity face reconstructions have so far been limited to either studio settings or through expensive 3D scanners. On the other hand, unconstrained reconstruction methods are typically limited by low-capacity models. Our method reconstructs accurate face geometry of a subject using a video shot from a smartphone in an unconstrained environment. Our approach takes advantage of recent advances in visual SLAM, keypoint detection, and object detection to improve accuracy and robustness. By not being constrained to a model subspace, our reconstructed meshes capture important details while being robust to noise and being topologically consistent. Our evaluations show that our method outperforms current single and multi-view baselines by a significant margin, both in terms of geometric accuracy and in capturing person-specific details important for making realistic looking models. 
\end{abstract}

\section{Introduction}

Reconstructing faces has been a problem of great interest in computer vision and graphics with applications in a wide variety of domains, ranging from animation \cite{ichim2015dynamic}, entertainment \cite{saito2016real}, genetics, bio-metrics, medical procedures, and more recently, augmented and virtual reality. Despite the long body of work, 3D face reconstruction still remains an open and challenging problem, primarily because of the high level of detail required owing to our sensitivity to facial features. Even slight anomalies in the reconstructions can make the output look unrealistic and hence, the accuracy of reconstructed face models is of utmost importance.

While accurate scans of facial geometry can be obtained using structured light or laser scanners, these are often prohibitively expensive, typically costing tens of thousands of dollars.
The seminal work of Beeler \etal \cite{beeler2010high} showed that a studio setup of cameras could be used to capture face geometry accurately. Since then, a variety of work has focused on using Photometric stereo or Multi-view stereo techniques in studio settings for face reconstruction and performance capture \cite{cao2018sparse, fyffe2017multi}. 
Although accurate in their reconstructions, these studio setups are not trivial to set up, typically requiring a calibrated camera setup along with controlled lighting and backgrounds. This makes them infeasible for capturing `in-the-wild' subject faces in unconstrained settings, for instance, an end user of a virtual reality app.

To tackle the problem of unconstrained 3D face reconstruction, the community has mostly relied on three-dimensional morphable models (3DMMs)~\cite{blanz1999morphable}. 3DMMs are low-dimensional linear sub-spaces of faces typically constructed using a small set of ground truth 3D scans that enable rapid approximation of face geometry, classically through a non-linear optimization over appearance and landmarks. Deep neural nets have more recently been used to fit morphable models using a single image. Generalization to in-the-wild images is often a concern for these methods. While the results are often visually appealing with texture, the reconstructions suffer from high geometric inaccuracies.

With the limited availability of 3D data for faces, using geometric cues from multiple views to improve the accuracy of reconstruction becomes necessary. Previous work has shown that a single template or 3DMM can be optimized using constraints from multiple views, using techniques like photometric stereo \cite{roth2015unconstrained} or advances in automatic keypoint detection \cite{huber2016multiresolution}.
Recently, Hernandez \etal \cite{hernandez2017accurate} proposed an elegant multi-view constrained structure-from-motion scheme that explicitly optimized the coefficients of a 3DMM shape to recover face geometry. However, the output still remains constrained to the underlying training data and low capacity of the 3DMM. This greatly limits its expressivity and is particularly undesirable for medical or bio-metric usage.

In this work, we attempt to answer the question ``What's the most accurate reconstruction an end-user can obtain, without needing access to special equipment or studio setups?". To this end, we propose a pipeline for highly accurate yet robust face geometry capture, requiring nothing but a smartphone. We leverage recent advances in the fields of object and keypoint detection, direct methods for visual SLAM, and higher frame-rate capture functionality available on modern smartphones. This allows us to incorporate multi-view consistency, landmark, edge and silhouette constraints into a single optimization framework. We also explicitly train a model for ear detection to incorporate ear landmarks, an area that has almost entirely been ignored in face reconstruction works. This enables us to achieve state-of-the-art geometric accuracy among unconstrained face reconstruction techniques. 

Our contributions are two-fold. First, we propose a 3D face reconstruction algorithm that takes a single video of a subject's face and reconstructs their face geometry, making high fidelity reconstructions accessible to users for downstream tasks like animation, printing, genetic analysis/modeling and biometrics. The reconstructed meshes have semantic correspondence and consistent topology, without being constrained to any model subspace.  Second, we release a dataset 200 video sequences of 100 individuals shot at 120fps, where we collect two sequences per subject, under varying lighting conditions. 


\section{Prior Work}
Prior work on 3D Face Reconstruction is substantially large and a complete literature review is out of scope for the paper. Here, we talk about about some of the works closely related to ours.

\noindent \textbf{SfM based Multi-view Reconstruction.} A lot of multi-view reconstruction methods employ a Structure-from-Motion pipeline ~\cite{gotardo2015photogeometric, lin2010accurate, fidaleo2007model} but with unconvincing results on unconstrained in-the-wild videos ~\cite{hernandez2017accurate}. ~\cite{brand2001morphable} and ~\cite{shi2014automatic} use 3D Morphable Model~\cite{blanz1999morphable} for fitting shapes on every frame after computing correspondences among them. This restricts the reconstruction to a low-dimensional linear subspace. The current state-of-the-art approach by Hernandez ~\etal ~\cite{hernandez2017accurate} uses 3DMM as a prior instead to search for correspondences among frames. This allowed them to achieve state-of-the-art results in unconstrained multi-view face reconstruction. However their method requires camera intrinsics to be known and the output is still constrained to a linear basis. We use this method as one of the baselines for comparison.

\noindent \textbf{Photometric Stereo.} Photometric stereo based methods have proven effective for large unconstrained collection of photos ~\cite{kemelmacher2011face, kemelmacher2013internet, roth2015unconstrained}. ~\cite{kemelmacher2011face} generates a 2.5D face surface by using SVD to find the low rank spherical harmonics. Roth \etal ~\cite{roth2015unconstrained} expand on it to handle pose variations and the scale ambiguity prevalent in the former method. They further expand their work in ~\cite{roth2016adaptive} where they fit a 3DMM to 2D landmarks for every image and optimize for the lighting parameters rather than SVD based factorization. Suwajanakorn ~\etal ~\cite{suwajanakorn2014total} use shape from shading coupled with 3D flow estimation to target uncalibrated video sequences. While these methods capture fine facial features, most of them rely on simplified lighting, illumination and reflectance models, resulting in specularities and unwanted facial features showing up on the mesh.

\noindent \textbf{Single Image 3D Face Reconstruction.} 3D Morphable Models have successfully been used as prior for modeling faces from a single image ~\cite{blanz1999morphable, breuer2008automatic, zhu2015high, saito2017photorealistic, jiang20183d, richardson20163d, tuan2017regressing}. Facial landmarks have commonly been used in conjunction with 3DMMs for the reconstruction ~\cite{zhu2015high, aldrian2010linear, kemelmacher20113d, dou2014robust}. While landmarks are informative for 3D reconstruction, relying primarily on them results in generic looking meshes which lack recognizable detail. More recently, convolutional neural networks have been put to use for directly regressing the parameters of the 3D Morphable Model ~\cite{zhu2016face, jourabloo2016large}. To overcome the limited expressivity of 3DMMs, recent methods have tried to reconstruct unrestricted geometry, by predicting  a volumetric representation ~\cite{jackson2017large}, UV map \cite{feng2018joint}, or depth map \cite{sela2017unrestricted}. However, the underlying training data of these methods has been limited to synthetic data generated using 3DMMs or course meshes fit using landmarks. Thus the ability of these methods to generalize to `in-the-wild' images and face geometry is still quite limited. While single image reconstruction is of great research interest, we believe multi-view consistency is crucial for generating accurate 3D face representations, specially given the limited data available for faces.  For a more comprehensive literature review of monocular 3D Face Reconstruction, we direct the readers to ~\cite{zollhofer2018state}.


\section{Method}
Our aim is to reconstruct a detailed 3D model of a subject's face, using a single video shot from a handheld device moving around the subject's face, while the subject is stationary. The video can be recorded by the subject themselves or by an assistant. 
Our approach takes advantage of the high frame rate capture available on modern smartphones, along with recent advances in automated keypoint and edge detection, to generate a highly accurate mesh of the face geometry. 
We process the video in three stages -  1) Camera pose estimation (\ref{sec:PBA}), 2) Point cloud generation using multi-view stereo (\ref{sec:pcl}) and 3) Mesh fitting using a combination of constraints. (\ref{sec:mesh_fit}).

\subsection{Camera pose estimation} \label{sec:PBA}
Most multi-view face reconstruction methods have traditionally relied on pre-calibrated cameras (a studio setup) or used landmark trackers for estimating camera pose relative to a geometric prior, such as a template mesh or 3DMM. However, landmark trackers are less than reliable beyond a small angle from the front of the face, which reduces their utility for camera pose estimation. For our method, we aim to get sub-pixel accurate camera pose estimates using recent advances in direct methods for visual SLAM, based on the seminal work by Engel \etal \cite{engel2018direct,engel2014lsd}. Direct methods are particularly effective for faces, where a lot of corner points are not present for feature point detection and matching.

We take advantage of the fact that the input is a single continuous video sequence. We use the geometric bundle adjustment based initialization scheme proposed in \cite{ham2017monocular} to get relative pose estimates for an initial baseline distance. Then, a LK tracker is used to track the camera frames in the video, and a keyframe is selected once the camera moves a certain baseline distance. The set of keyframe camera poses are optimized using photometric bundle adjustment to maximize photometric consistency between frames.

As in \cite{engel2018direct}, PBA is a joint optimization of all model parameters, including camera poses, the intrinsics, and the radial distortion parameters. For a typical sequence, 50-80 keyframes with accurately known camera poses are obtained.

Independently of pose estimation, we use the publicly available Openface toolkit \cite{baltrusaitis2018openface} for facial landmark detection. We fit the Basel 3DMM \cite{blanz1999morphable} to these landmarks and align it with the coordinate system of the keyframes. We use this coarse mesh in the next stage.

The advantages of decoupling camera pose estimation and face alignment are three-fold: 1) Robustness to landmark tracker failures, which, despite many recent advances, is not robust at large angles  2) By not relying on the estimated coarse mesh for registering camera poses, errors in the shape estimation do not propagate to the camera poses. 3) Purely using photometric consistency allows us to achieve sub-pixel accuracy in estimating camera poses. 

\subsection{Point cloud generation} \label{sec:pcl}

At the end of the PBA stage, we obtain a set of 50-80 keyframes whose camera poses are known with high accuracy, and a coarse face mesh fitted to the landmarks from Openface. Next, we use these keyframes to generate a dense point cloud of the face geometry using Multi-view stereo.
 We use the parallelized multi-view PatchMatch implementation of Galliani \etal \cite{galliani2015massively} and use 12 source views for each reference view for depth inference. The multi-view PatchMatch estimates a depth map for each of the keyframes. We initialize the depths and search range using the coarse mesh.

To select which source views to use to infer the depth map of a reference view, we calculate a view selection score \cite{yao2018mvsnet} for each pair of keyframes, $s(i, j) = \sum_{\mathbf{p}} \mathcal{G}(\theta_{ij}(\mathbf{p}))$ , where $\mathbf{p}$ is a point common to both views and its baseline angle between the cameras $\mathbf{c}_i$ and $\mathbf{c}_j$ is $\theta_{ij}(\mathbf{p}) = (180/\pi)\arccos((\mathbf{c}_i - \mathbf{p}) \cdot (\mathbf{c}_j - \mathbf{p}))$.  $\mathcal{G}$ is a piecewise Gaussian, as follows :
\[ \mathcal{G}(\theta) =  \left\{
\begin{array}{ll}
      \exp(-\frac{(\theta - \theta_0)^2}{2\sigma_1^2}), \theta \leq \theta_0 \\
      \exp(-\frac{(\theta - \theta_0)^2}{2\sigma_2^2}), \theta > \theta_0 \\
\end{array} 
\right. \]
For our method, we pick $\theta_{0} =10 $  , $\sigma_{1} =5$  and $\sigma_2=10$  .
 We use the estimated coarse mesh to filter out noisy patches in the depth maps produced by the PatchMatch. We then project the depth maps to a single fused point cloud using the fusion strategy proposed in \cite{galliani2015massively}.

\begin{figure}[t]
\begin{center}
   \includegraphics[width=0.95\linewidth]{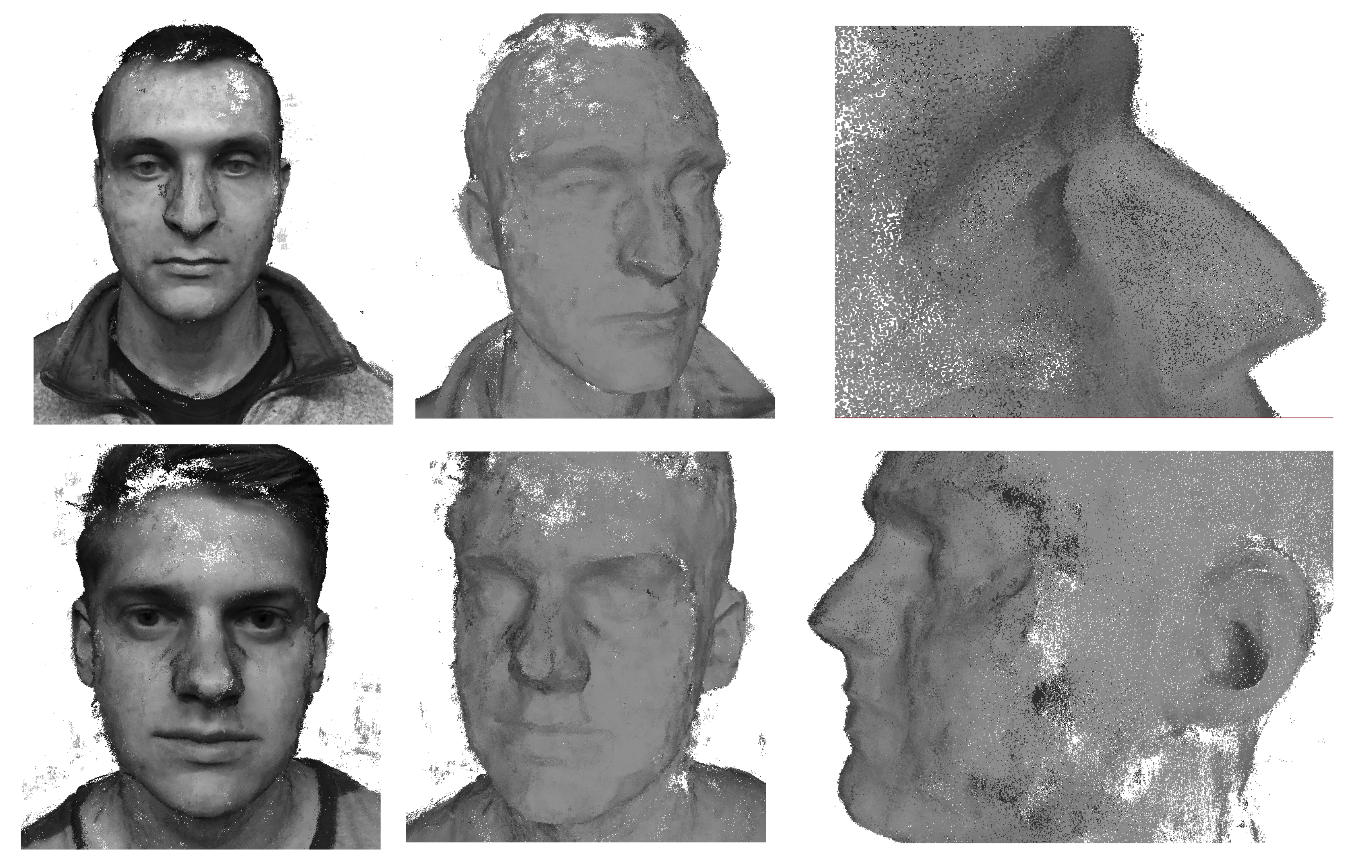}
\end{center}
   \caption{Example point clouds generated at the end of our Point cloud generation stage, with and without texture. The point clouds accurately capture the overall face geometry and details in areas like eyes and lips, that make the person recognizable. However, the point clouds have missing data as well as noise, which requires a robust mesh fitting approach (\ref{sec:mesh_fit}).}
\label{fig:pcl_sample}
\end{figure}

Example point clouds output at this step are visualized in Fig \ref{fig:pcl_sample}.

\subsection{Mesh fitting} \label{sec:mesh_fit}

\begin{figure}[t]
\begin{center}
   \includegraphics[width=0.95\linewidth]{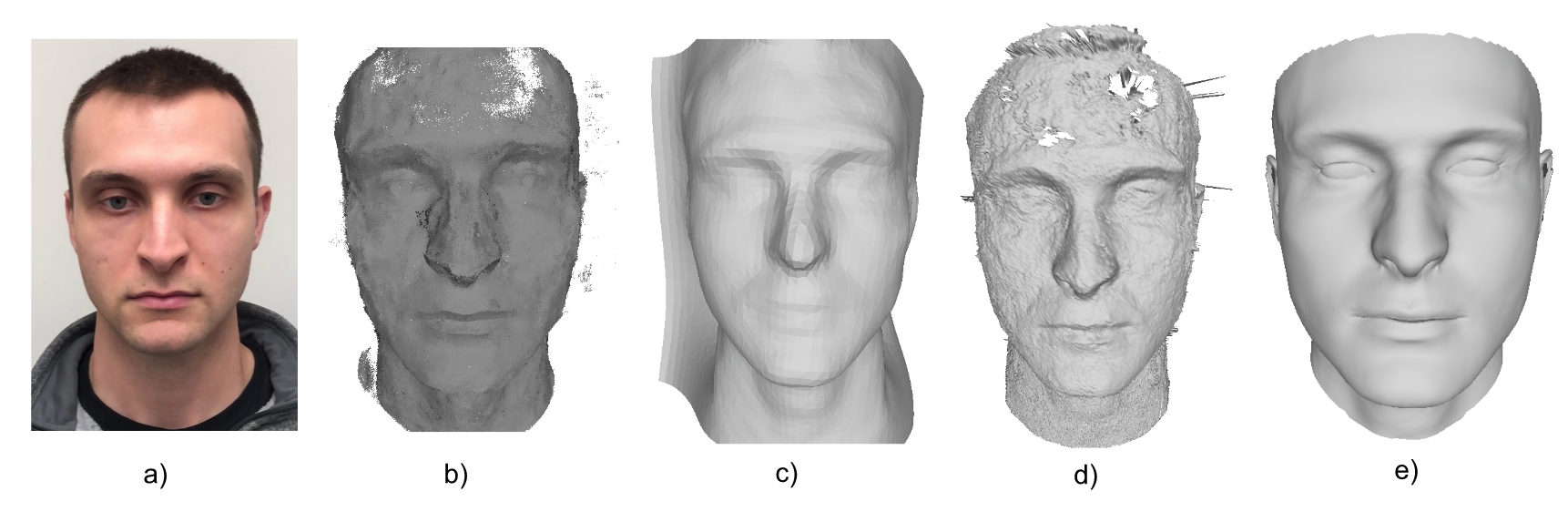}
\end{center}
   \caption{Comparison of mesh generation methods a) Sample image. b) Generated point cloud (\ref{sec:pcl}) c) \cite{kazhdan2013screened} can fill in gaps in the point cloud but at the cost of overly smooth meshes. d) Depth fusion method of \cite{hernandez2015near} can preserve details, but is unable to handle missing data. e) Our approach reconstructs meshes with consistent topology and correspondence between vertices, while capturing details of the point cloud and being robust to noise and missing data. }
\label{fig:mesh_comp}
\end{figure}

Due to factors like non-ideal lighting, lack of texture and sensor noise of the smartphone, the obtained point cloud typically has noise and incompletions, with the points distributed around the `true' surface.
 Techniques like Screened Poisson reconstruction or the depth map fusion strategy of \cite{hernandez2015near} either return meshes with a lot of surface noise or extremely smoothed out details, depending on the regularization used (see Fig. \ref{fig:mesh_comp}). Further, for the reconstructed mesh to be of use in further downstream tasks such as animation, bio-metrics or as input to a learning algorithm, it is extremely desirable for the meshes to have a consistent topology. 

 Statistical ICP inspired techniques have proposed fitting a 3DMM to a point cloud \cite{schneider2009fitting,bazik2017robust,blanz2004statistical} in the past. However, fitting a 3DMM defeats the purpose of not being constrained to an existing linear basis of shape.
 We adapt non-rigid mesh fitting algorithm \cite{amberg2007optimal}, originally proposed for registering template meshes to 3D scanner data, to deform a template using a combination of constraints given by the point cloud, landmarks, mesh stiffness and edge constraints.

 \subsubsection{Point cloud constraints}
 The primary constraint for the mesh deformation comes from the 3D information captured in the point cloud.
 While well-studied techniques exist to register a template mesh to a 3D scanned mesh \cite{amberg2007optimal}, registering a mesh to point clouds of the sort obtained from multi-view stereo techniques is more challenging. For example, simply fitting each vertex to its nearest-neighbor in the point cloud will cause the mesh to become extremely noisy, as there will be many outlier points.
 
\begin{figure}[t]
\begin{center}
   \includegraphics[width=0.5\linewidth]{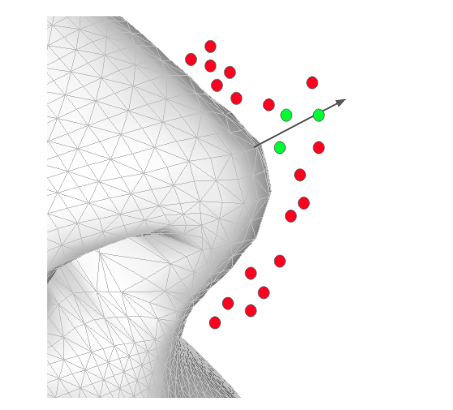}
\end{center}
   \caption{Exaggerated view of the point cloud constraints. For each vertex, the set of points within a small threshold of its normal (in green here) are found and their median used as the target 3D coordinate for the vertex. }
\label{fig:mesh_fit_pcl}
\end{figure}

To address this, we take advantage of the fact that for a template mesh, the vertex normals can be easily estimated. For each vertex, we select the points in its neighborhood, and for each point, we calculate its perpendicular distance to the normal of the vertex. Points within a small threshold distance are accepted while the rest are rejected (see Fig. \ref{fig:mesh_fit_pcl}. 
 
 For each vertex on the template mesh we obtain its desired location in 3D as the median of the accepted points.
 
 \subsubsection{Landmark constraints}

 The second source of information are the 68 2D landmarks obtained using the automatic landmarking solution of \cite{baltrusaitis2018openface}. Landmarks are important for global alignment and scaling of the mesh, as well as ensuring all the reconstructed meshes are in semantic correspondence.
 
 For the set of frames for which the landmarks have been annotated with high confidence by the tracker (typically close to frontal poses), we solve for the 3D locations of the landmarks by minimizing geometric reprojection error, 
 \begin{equation}
    E_{X_{j}} = \sum_{i} \sum_{j} d(\pi (\theta_{i},X_{j}),x_{ij})^2
\end{equation}
Where $\theta_i$ is the $i$-th camera's pose, $X_{j}$ is the $j$-th landmark's coordinates in 3D, and $x_{ij}$ is the 2D coordinate of the landmark returned by the landmark tracker for the $i$-th frame. For our purposes, we ignore the 18 landmarks corresponding to the face contour, and use the remaining 50 landmarks as constraints for the corresponding 3D vertices.

Historically, most landmark trackers have focused only on these 68 keypoints. As a consequence, many reconstruction techniques either focus only on reconstructing the frontal face region, or generate a full mesh but evaluate only on the frontal section. Ears and the side facial regions have mostly been ignored in previous works. Even learning-based face alignment techniques do not do well on the ears, as the underlying training data is based on the annotation/detection of these 68 landmarks.

\begin{figure}[t]
\begin{center}
   \includegraphics[width=0.7\linewidth]{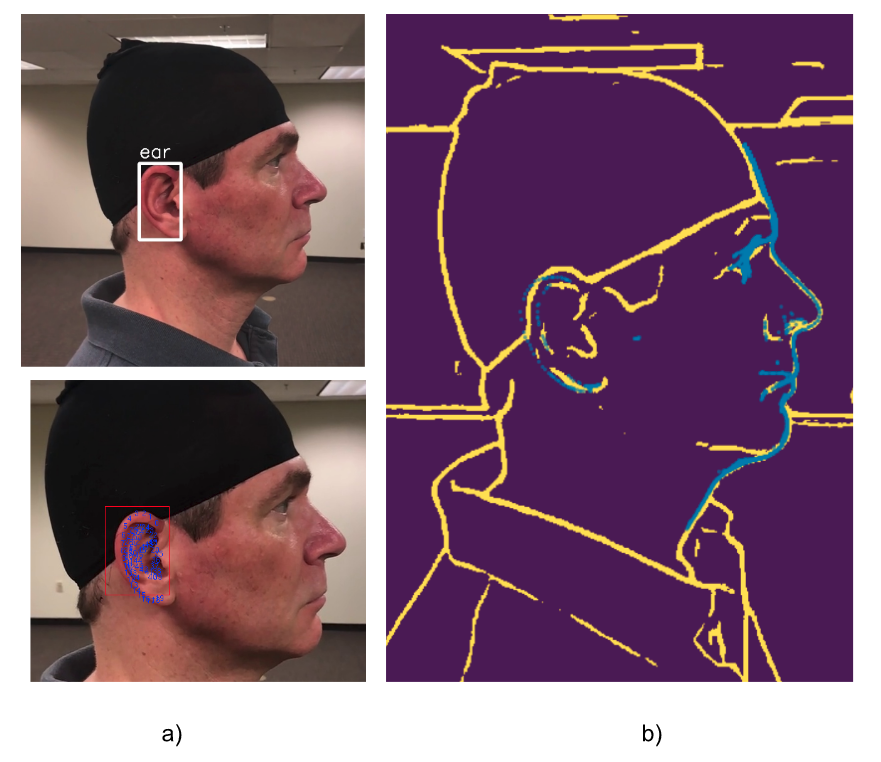}
\end{center}
   \caption{ a) We train a bounding box regressor (above) and landmark detector (below) specifically for ears. This improves our reconstruction's overall accuracy while allowing us to capture ear size and contour. b) Visualization of edge constraints. Image edges in yellow, mesh vertices corresponding to edges projected in blue. Note that mesh vertices fit the ear well because of the ear landmark detection. }
\label{fig:ear_lm_and_edges}
\end{figure}

To explicitly address this, we make use of a recent dataset of `in-the-wild' ear images annotated with bounding boxes and landmarks \cite{zhou2017deformable}. We first train the deep object detection model of Redmon \etal \cite{redmon2017yolo9000} for a single `ear' class. We then train an ensemble of regression trees \cite{kazemi2014one} for predicting landmarks using the bounding box detection as input. As seen in Fig \ref{fig:ear_lm_and_edges}, despite the limited training data size, we are able to achieve impressive robustness and accuracy in the landmark detection. We use a subset of the landmarks corresponding to the outer contour of the ear as additional landmark constraints in our mesh fitting. To the best of our knowledge, ours is the first face reconstruction method to explicitly address the ears, which in turn improves overall accuracy and metrics like the width of the face.

\subsubsection{Edge constraints}
Silhouette constraints have shown to be powerful cues in recent 3D reconstruction literature \cite{alldieck2018detailed,bas2016fitting}. For faces, views that are close to profile are particularly informative. However, since many single and multi-view approaches rely on landmarking for camera pose estimation, they fail to make use of silhouettes beyond a certain angle. By solving for the camera poses independently of landmarking, we can actually make use of extreme profile views. This proves to be helpful in capturing challenging areas for face reconstruction algorithms, such as the nose, lips and lower chin/neck region.
We use a combination of Z-buffering \cite{Foley1990ComputerG} and backface-culling  to estimate vertices that project an edge onto a given view. To find the corresponding edges in the RGB image, we use the Structured Forests edge detection approach proposed in \cite{dollar2013structured}. For each vertex projecting an edge in the frame, its nearest neighbor is found in the edge map. This corresponding point is back-projected in 3D to obtain a `target' location for the vertex in 3D.

 \subsubsection{Putting it all together}
 With the combination of the cues from the point cloud, landmarks, and silhouettes, we obtain a set of constraints that we wish to use to deform the template mesh. For a template mesh M of fixed topology (V, E), this can be written as a weighted sum of energies we wish to minimize: 
 $$ \arg \min_{V}  E_{pcl} + \alpha E_{lms} + \beta E_{edges} +  \gamma E_{reg} $$
where $E_{reg}$ is a regularization energy arising from the mesh topology that restricts connected vertices to deform similarly. This system can naturally be expressed in the iterative linear system-based non-rigid mesh registration algorithm proposed by Amberg \etal \cite{amberg2007optimal}. 
 
 At each iteration, a linear system of the form $\mathbf{A}\mathbf{X} = \mathbf{B}$ is solved, where $\mathbf{X}$ is a $4n\times3$ matrix, containing the per-vertex 3x4 affine transform matrix. The matrix $\mathbf{A}$ captures information of the source template in terms of the mesh connectivity and vertex locations. The mesh connectivity acts as an adjustable `stiffness' regularization, which controls how much neighboring vertices can move with respect to each other. The matrix 
 $\mathbf{B}$ contains the corresponding `target' locations in 3D, such as those obtained from the point cloud, landmarks and edges. The mesh stiffness and the weights of the landmarks are gradually decreased, gradually moving from global stretching to local, data-driven deformations. After every few iterations, the point cloud and edge constraints are recalculated using the current locations of the vertices. For further details, we refer the reader to the original paper \cite{amberg2007optimal}. For our template, we use the Basel 3DMM mesh \cite{blanz1999morphable}, simply because of its prevalence as an input or output of several face reconstruction algorithms.



\begin{figure*}
\begin{center}
   \includegraphics[width=0.95\linewidth]{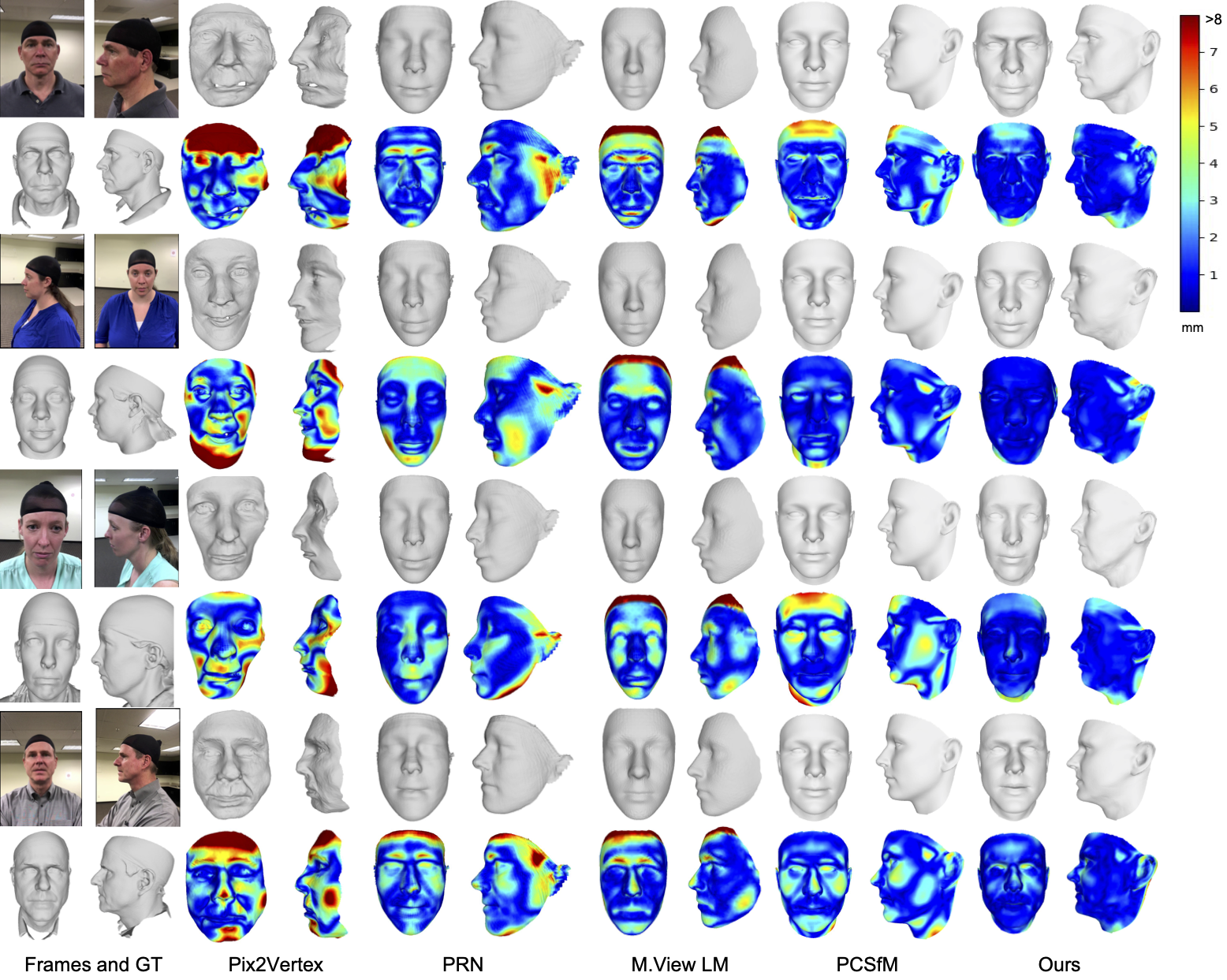}
\end{center}
  \caption{Qualitative comparison against reconstructions of various single and multi-view methods. Let to Right: sample frame and ground truth 3D, Pix2vertex \cite{sela2017unrestricted}, PRN \cite{feng2018joint}, multi-view landmark fitting (4dface \cite{huber2016multiresolution}), PCSfM \cite{hernandez2017accurate}, Ours. For each method the upper row shows the reconstructed mesh, front and profile, and the corresponding heatmap of error (Accuracy) is shown in the lower row }
\label{fig:results}
\end{figure*}

\section{Results}

For evaluating our approach, we collect a dataset of videos using an iPhone X, with a stationary subject holding a static expression and the camera moving from one profile to the other. The videos are shot at the 120fps setting and are typically 15-20 seconds long, depending on whether they are shot by the subject themselves or by an assistant. The background conditions are unconstrained, though we do ensure a mostly static scene to get accurate camera pose estimates from our Photometric Bundle Adjustment step. The subject maintains a static expression and occlusions from hair are avoided when needed using caps. Our method takes 30-40 minutes of processing time, with the MVS step the most computationally expensive.

\subsection{Quantitative Evaluation} \label{sec:quant}

For 10 subjects among the videos we collected, we obtained high accuracy 3D face scans using an Arctic Eva structured light hand-held scanner. The scans were obtained immediately after the video was recorded with the subjects still in the same poses, to ensure no discrepancy in face geometry between the video and the scan. We use the videos of these subjects to reconstruct face meshes using the methods listed in Table \ref{table:results}. For methods that work on a single image, we use a close to frontal keyframe as input. For the edge-fitting based single view method of Bas \etal \cite{bas2016fitting}, we select a frame at roughly 15 degrees to the front, as reported in their paper. For the multi-view methods, we either use the keyframes generated by our method or the whole video, depending on what the method uses as input.
For all methods except PCSfM, the authors make their implementations public and we use those for evaluation. For PCSfM, we use our own implementation.

A challenge in fair quantitative evaluation arises from the fact that different methods reconstruct different amounts of the face area as defined by the Basel mesh, such as frontal only in pix2vertex \cite{sela2017unrestricted} , front and side without ears in PRN \cite{feng2018joint}, full Basel mesh for PCSfM \cite{hernandez2017accurate} and arbitrary in SfM (using COLMAP \cite{schonberger2016structure}). To address this, we first register a standard Basel mesh to the ground truth 3D scans using Non-rigid ICP \cite{amberg2007optimal,booth2018large}. We then borrow a strategy from MVS benchmarks \cite{jensen2014large,knapitsch2017tanks,yao2018mvsnet} to evaluate the reconstructions using \textbf{Accuracy} - the distance from the reconstruction's vertices to the ground truth, and \textbf{Completion} - the distance from the ground truth's vertices to the reconstruction. Thus, if a method reconstructs only the frontal face, it might do well in Accuracy and be penalized in Completion. We report the mean and median of these distances, averaged over the 10 subjects, in Table \ref{table:results}.

We compare our methods against several recent single and multi-view reconstruction methods. 
As can be observed, single view methods typically have very poor performance in terms of accuracy and completion. 
As also noted in \cite{hernandez2017accurate}, certain methods that just reconstruct smooth meshes tend to have low numeric errors, even if the reconstruction lacks details important for making a person recognizable.

Our method clearly outperforms single and multi-view baselines, both in terms of accuracy and completion. We note that our median accuracy is around 0.95 mm, showing that for majority of the mesh we achieve sub-millimeter accuracy.

\textbf{Ablation.} We generate reconstructions without Edge constraints and without the ear landmarks respectively. Edges improve the accuracy of the reconstructions by improving the fit of areas like the jaw and nose, whereas the ear landmarks improve the information captured in the ears as well as overall mesh scale and width. Thus dropping either leads to a drop in accuracy. A significant drop in completion is also observed when removing the ear landmarking, because the reconstructed mesh moves away from the ears of the ground truth mesh.


\begin{table*}
\begin{center}
\begin{tabular}{|c|c|c c c|c c c|}
\hline 
\multirow{2}{*}{Method} & \multirow{2}{*}{Views} 
& \multicolumn{3}{c|}{Accuracy (mm) } 
& \multicolumn{3}{|c|}{Completion(mm) } \\ 
\cline{3-8}
& & Mean & Std. Dev. & Median & Mean & Std. Dev. & Median\\
\hline\hline
Mean Basel \cite{blanz1999morphable}       & - & 3.09  & 1.24 &  2.62 & 3.02 & 1.25  & 2.76 \\ 
Landmark fitting \cite{huber2016multiresolution}  & Single  & 2.53  & 0.62 & 1.88 & 8.01     & 2.13 & 3.62      \\ 
pix2vertex \cite{sela2017unrestricted} & Single & 3.49 & 0.76 & 2.76 & 25.33 & 4.62 & 16.34  \\
PRN \cite{feng2018joint} & Single & 2.63 & 0.84 & 2.30 & 6.27 & 2.17 & 3.24 \\
Edge-fitting \cite{bas2016fitting} & Single & 3.06 & 1.28 & 2.63 & 3.02 & 1.25 & 2.75 \\

\hline
M.view lm fit \cite{huber2016multiresolution,huber2015fitting} & Multi   & 2.23 & 0.41 &    1.69   & 7.87          & 1.98 & 3.59   \\ 
Roth \etal \cite{roth2015unconstrained}     & Multi   & 3.31 & 1.03 & 2.65 & 7.65 & 1.86 & 3.67 \\
SfM \cite{schonberger2016structure} & Multi & 5.42 & 2.55 & 3.61 & 4.72 & 2.41 & 3.60 \\
PCSfM* \cite{hernandez2017accurate} & Multi & 1.87 & 0.40 & 1.66 & 2.38 & 0.85 & 2.04 \\
\hline
Ours w/o Edges    &  Multi  & 1.38 & 0.24 & 0.98 & 1.30 & 0.29 & 0.97 \\
Ours w/o ear lms    &  Multi     & 1.33 & 0.27 & 0.96 & 1.41 & 0.36& 1.07      \\
Ours    &  Multi     & \textbf{1.24} & 0.26 & \textbf{0.95} & \textbf{1.29} & 0.29 & \textbf{0.95}     \\

\hline
\end{tabular}
\end{center}
\caption{Quantitative results against ground truth scans. We evaluate the state of the art single and multi-view reconstruction methods. As is common in MVS benchmarks, we evaluate the reconstructions in terms of average distance from reconstruction to ground truth (accuracy) and distance from ground truth to reconstruction (completion). All numbers in mm; lower is better. * denotes that the method needs camera intrinsics to be known in advance.}
\label{table:results}
\end{table*}

\subsection{Qualitative evaluation} \label{sec:qual}

Fig. \ref{fig:results} compares our mesh reconstructions with the existing single and multi-view face reconstruction methods. We can clearly observe the limitations of single-view methods ~\cite{sela2017unrestricted, feng2018joint} in capturing an accurate face geometry. Meshes from ~\cite{huber2016multiresolution} are overly smooth. Except ours, all methods are unable to capture the profile geometry of faces. 

\begin{figure}
\begin{center}
   \includegraphics[width=0.6\linewidth]{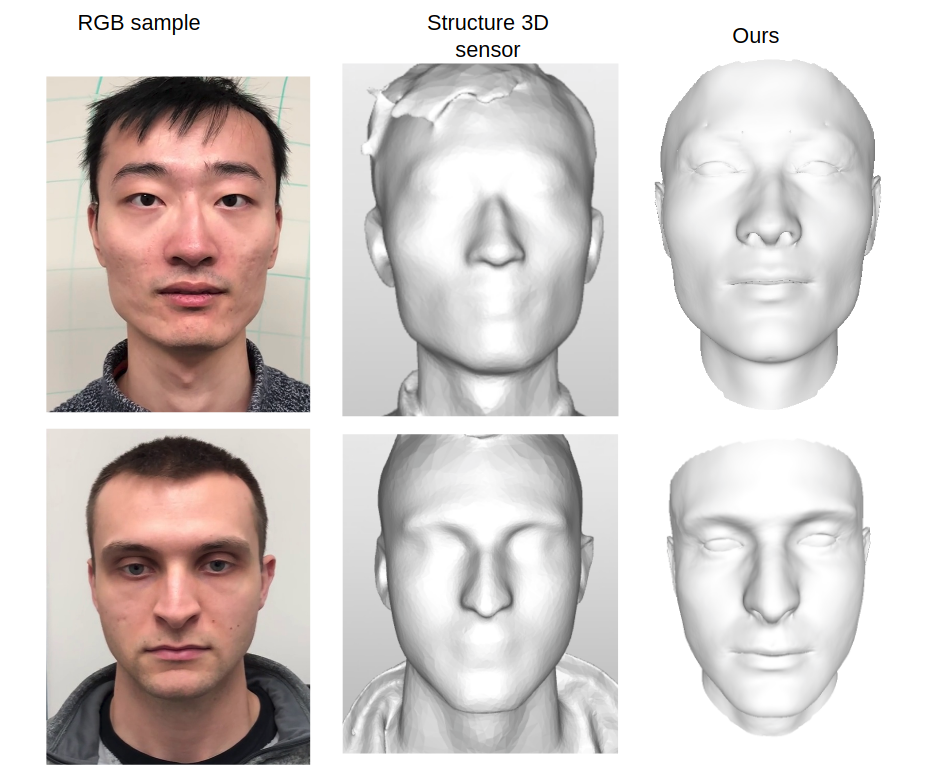}
\end{center}
   \caption{(Middle) Output from Structure RGB-D Sensor \cite{structure2019}. Details like the eyes, nose and lips are excessively smoothed out. (Right) Our reconstruction.}
\label{fig:long}
\label{fig:onecol}
\end{figure}

Fig. \ref{fig:long} shows a comparison between our results against against scans obtained by a commercially available IR-based RGB-D sensor ~\cite{structure2019}. We can notice that our method preserves a lot more facial details while being relatively more accessible to an end user.

\begin{figure}
\begin{center}
   \includegraphics[width=0.5\linewidth]{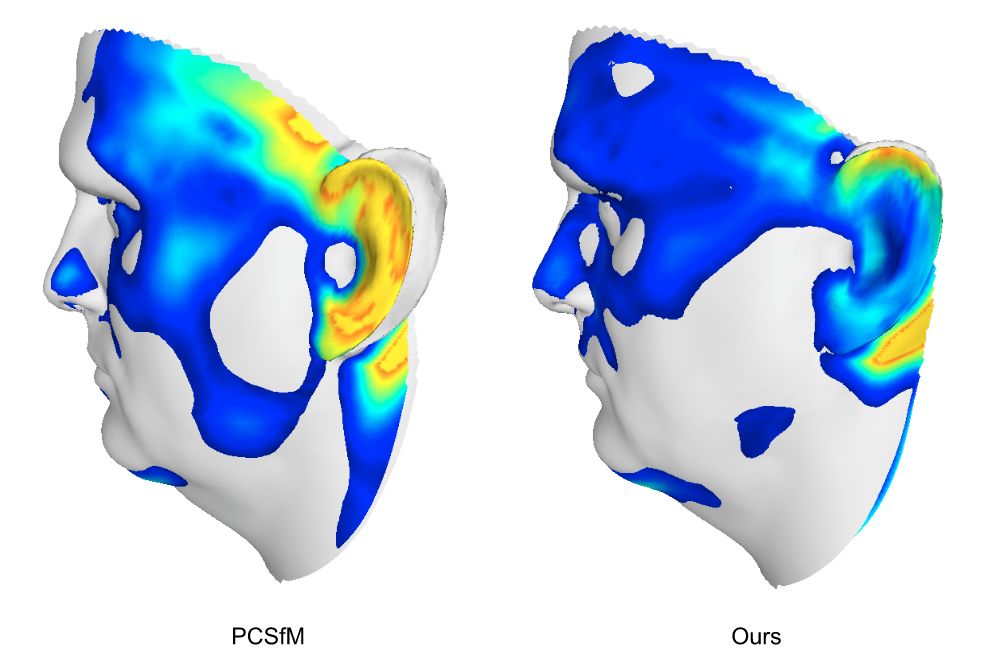}
\end{center}
   \caption{Effect of ear landmarking: Ground truth mesh (white) overlapped with error heatmaps of PCSfM(left) and ours(right). Landmarking the ears greatly improves our fitting and reduces the geometric error in our reconstructions }
\label{fig:ear_align}
\end{figure}





\textbf{Fine detail enhancement}. A recent trend in the 3D face reconstruction research has been to emboss fine high-frequency details to the reconstruction with methods like shape-from-shading \cite{or2015rgbd} or mesoscopic augmentations \cite{beeler2010high}. While not always reflecting the true geometry of the surface, these methods add realism to the mesh, which can be desirable for purposes like animation. Such methods can easily be applied to our reconstructions as well. We modify the mesoscopic augmentation method proposed in \cite{sela2017unrestricted} so that the underlying geometry is preserved, and apply it to our meshes. Since these are based on a dark-is-deep assumption, we skip quantitative evaluation, and provide qualitative results in Fig.\ref{fig:high_freq}. Details on the modified approach are provided in the supplementary.

 \begin{figure}
\begin{center}
  \includegraphics[width=0.6\linewidth]{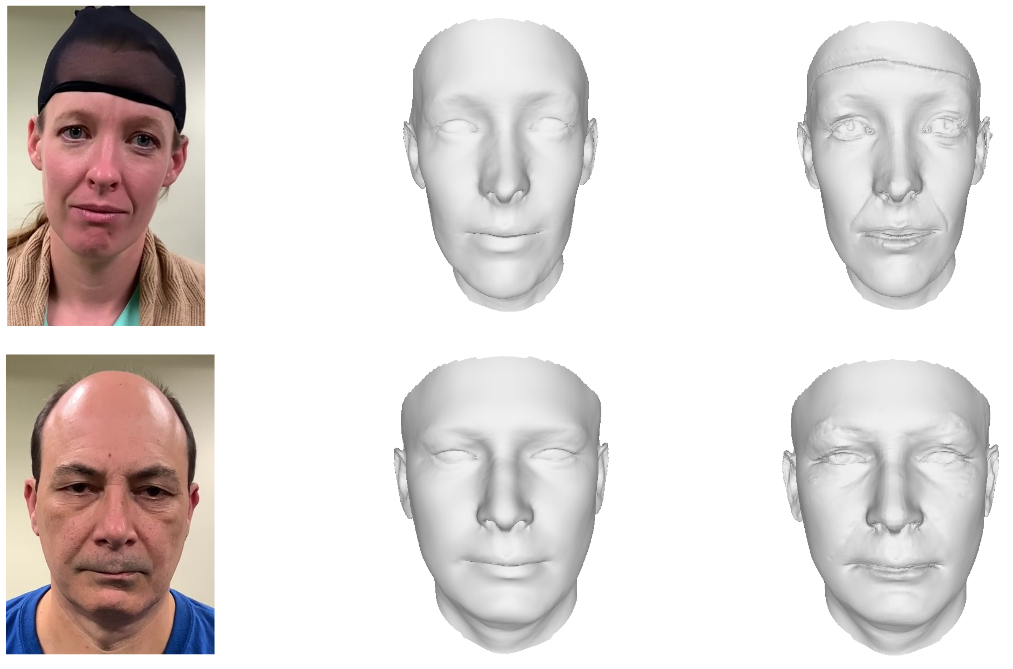}
\end{center}
  \caption{(Centre) Ours. (Right) Ours with modified mesoscopic augmentations.}
\label{fig:high_freq}
\end{figure}


\section{Dataset}

Our results reaffirm that incorporating multi-view consistency in 3D reconstruction greatly improves the quality and reliability of the results. Incorporating geometric structural priors into deep learning based reconstruction has shown to be extremely effective \cite{yao2018mvsnet}, even with moderate amounts of training data. The dearth of multi-view data for faces (see Table \ref{tab:datasets}) has prohibited progress in this space. We make our dataset of 100 subjects available, with 2 video sequences recorded per subject under different lighting and background conditions. For each video, we provide a set of 50-80 keyframes we used and our reconstructions (mesh, point clouds and surface normal maps) for reference. For a subset of the data we also provide high accuracy meshes obtained using a structured light scanner. For the rest of the scans, for each subject we validate the meshes to be self-consistent between the two sequences, within a small tolerance. More details on this are covered in the supplementary.
We hope that this dataset will help further the research and evaluation of unconstrained multi and single view reconstruction algorithms that should be both accurate and consistent. It will especially enable self-supervised methods that enforce consistency across views and between sequences.

\begin{table}
\begin{center}
\begin{tabular}{|l|c| c|}
\hline
Dataset & Subjects &  Poses \\
\hline\hline
ND-2006 & 888 & None \\
BU-3DFE & 100 & 2 \\
Texas 3DFRD & 118 & None\\
Bosphorus & 105 & 13\\
CASIA-3D & 123 & 11\\
MICC & 53 & 3\\
UHDB11 & 23 & 12\\
\hline
\end{tabular}
\end{center}
\caption{An overview of available 3D face datasets and the pose variation in RGB images available in them.}
\label{tab:datasets}
\end{table}

\section{Conclusion and Future Work}
In this work, we present a practical solution for an end user to capture accurate face geometry without using any specialized sensor. We improve over the prior work in several aspects: Our optimization scheme allows integration of landmark, edge and point cloud constraints from multiple frames. Experiments demonstrate better face reconstructions, both quantitatively and qualitatively. 
Since we optimize over an unrestricted geometry, our method is slower than many recent learning based methods. Further, our PBA based pose estimation is not robust to dynamic movements in the scene. Deep learning methods have proven to be effective in overcoming these shortcomings but this has not translated to face reconstruction research due to lack of data. We plan to address this in our future work and hope that our proposed pipeline and dataset will further research in this direction.    

\newpage
\newpage
{\small
\bibliographystyle{ieee}
\bibliography{egbib}
}

\end{document}